\documentclass[letterpaper, 10 pt, conference]{ieeeconf}  

\IEEEoverridecommandlockouts                              

\usepackage{graphics} 
\usepackage{epsfig} 
\usepackage{mathptmx} 
\usepackage{times} 
\usepackage{amsmath} 
\usepackage{amssymb}  

\usepackage{cite} 
\usepackage{url} 
\usepackage{booktabs}  
\usepackage{graphicx} 
\usepackage{dblfloatfix}
\usepackage{balance}  

\usepackage{multirow}
\usepackage{tikz}

\title{\LARGE \bf
 On Learning Spatial Structure from Pre-Beamforming Per-Antenna Range-Doppler Radar Measurements
}

\author{George Sebastian, Philipp Berthold, Bianca Forkel, Leon Pohl, and Mirko Maehlisch
\thanks{This work was supported in part by the Federal Office of Bundeswehr Equipment, Information Technology and In-Service Support (BAAINBw) and in part by dtec.bw – Digitalization and Technology Research Center of the Bundeswehr (project MORE), funded by the European Union – NextGenerationEU. We acknowledge financial support by University of the Bundeswehr Munich.
}
\thanks{The authors are with the Institute for Autonomous Driving, Department of Aerospace Engineering, University of the Bundeswehr Munich, Neubiberg, Germany. Corresponding author:
        {\tt george.sebastian@unibw.de}}%
}

\begin{document}

\maketitle

\begin{tikzpicture}[remember picture,overlay]
\node[
    anchor=north west,
    font=\footnotesize
] at ([xshift=0.55in,yshift=-0.28in]current page.north west)
{IEEE Robotics and Automation Letters (RA-L), 2026. DOI: 10.1109/LRA.2026.3739028};
\end{tikzpicture}

\begin{tikzpicture}[remember picture,overlay]
\node[
    anchor=south west,
    text width=7.4in,
    align=left,
    font=\scriptsize
] at ([xshift=0.55in,yshift=0.20in]current page.south west)
{\textcopyright~2026 IEEE. Personal use of this material is permitted.
Permission from IEEE must be obtained for all other uses, in any current or future media,
including reprinting/republishing this material for advertising or promotional purposes,
creating new collective works, for resale or redistribution to servers or lists,
or reuse of any copyrighted component of this work in other works.};
\end{tikzpicture}

\thispagestyle{empty}
\pagestyle{empty}

\begin{abstract}
Automotive radar perception pipelines commonly construct angle-domain representations via beamforming before applying learning-based models. This work instead investigates a representational question: can meaningful spatial structure be learned directly from pre-beamforming per-antenna range-Doppler (RD) measurements? Experiments are conducted on a 6-TX $\times$ 8-RX (48 virtual antennas) commodity automotive radar employing an A/B chirp-sequence frequency-modulated continuous-wave (CS-FMCW) transmit scheme, in which the effective transmit aperture varies between chirps (single-TX vs.\ multi-TX), enabling controlled analyses of chirp-dependent transmit configurations. We operate on pre-beamforming per-antenna RD tensors using a dual-chirp shared-weight encoder trained in an end-to-end, fully data-driven manner, and evaluate spatial recoverability using bird's-eye-view (BEV) occupancy as a geometric probe rather than a performance-driven objective. Supervision is visibility-aware and cross-modal, derived from LiDAR with explicit modeling of the radar field-of-view and occlusion-aware LiDAR observability via ray-based visibility. Through analyses of signal properties, transmit configurations (A-only, B-only, and A+B), receive aperture, and range-Doppler structure, together with physics-aligned baselines, we investigate the factors influencing spatial recoverability. The results indicate that meaningful spatial structure is recoverable from pre-beamforming per-antenna RD tensors under the studied A/B CS-FMCW radar configuration through learned spatial mixing, without relying on hand-crafted signal-processing stages.

\end{abstract}

\section{INTRODUCTION}

Automotive radar perception pipelines commonly recover spatial structure through beamforming or angle FFT, often followed by hand-crafted signal-processing stages such as constant false alarm rate (CFAR) detection, before applying learning-based models~\cite{10952908}. In practice, learning-based methods typically operate on angle-resolved representations such as range-azimuth (RA) maps, range-azimuth-Doppler (RAD) tensors, 4D radar cubes, or radar point clouds~\cite{9022248, Huang_2025_ICCV, 9469418, 9353210, scheiner_object_2021, Fent_2023_CVPR}. This separation reflects the conventional design choice that spatial mixing is performed prior to learning-based perception.

\begin{figure}[t]
\centering
\includegraphics[width=\columnwidth]{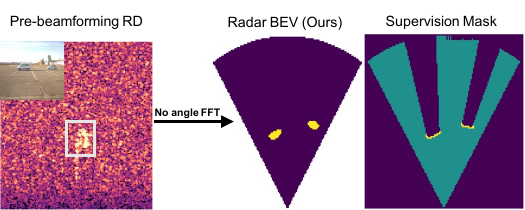}
\caption{Despite ambiguous RD observations (near-zero Doppler), meaningful spatial structure is recoverable in BEV through learned spatial mixing.}
\label{fig:teaser}
\end{figure}

However, spatial information is carried by the inter-antenna phase relationships before angle-domain construction~\cite{7870764, 8830483,11654526}. This raises a representational question: is explicit angle-domain processing necessary, or can spatial structure relevant for geometric reasoning be learned directly from pre-beamforming per-antenna range-Doppler (RD) measurements? As illustrated in Fig.~\ref{fig:teaser}, even ambiguous RD observations can support recovery of meaningful spatial structure in BEV through learning. 

Although prior work has demonstrated learning from RD representations~\cite{9879148, 9827281, 11127345, 10994166} or raw radar signals~\cite{ADCNet, 11130063}, these approaches are typically evaluated through downstream perception tasks such as detection, tracking, or free-space segmentation, which are optimized for task performance rather than explicitly probing geometric recoverability. In contrast, we study whether spatial geometry can be recovered directly from per-antenna RD measurements, using bird's-eye-view (BEV) occupancy as a geometric probe task over the observable scene.

We study this question using a 6-TX $\times$ 8-RX automotive radar employing an A/B chirp-sequence frequency-modulated continuous-wave (CS-FMCW) waveform~\cite{smartmicro_drvegrd152}, in which different chirps activate different transmit configurations (single-TX vs.\ multi-TX). This results in chirp-dependent transmit apertures, enabling controlled analyses of how transmit configuration influences spatial recoverability.

\begin{figure*}[t]
    \centering
    \includegraphics[width=\textwidth, keepaspectratio]{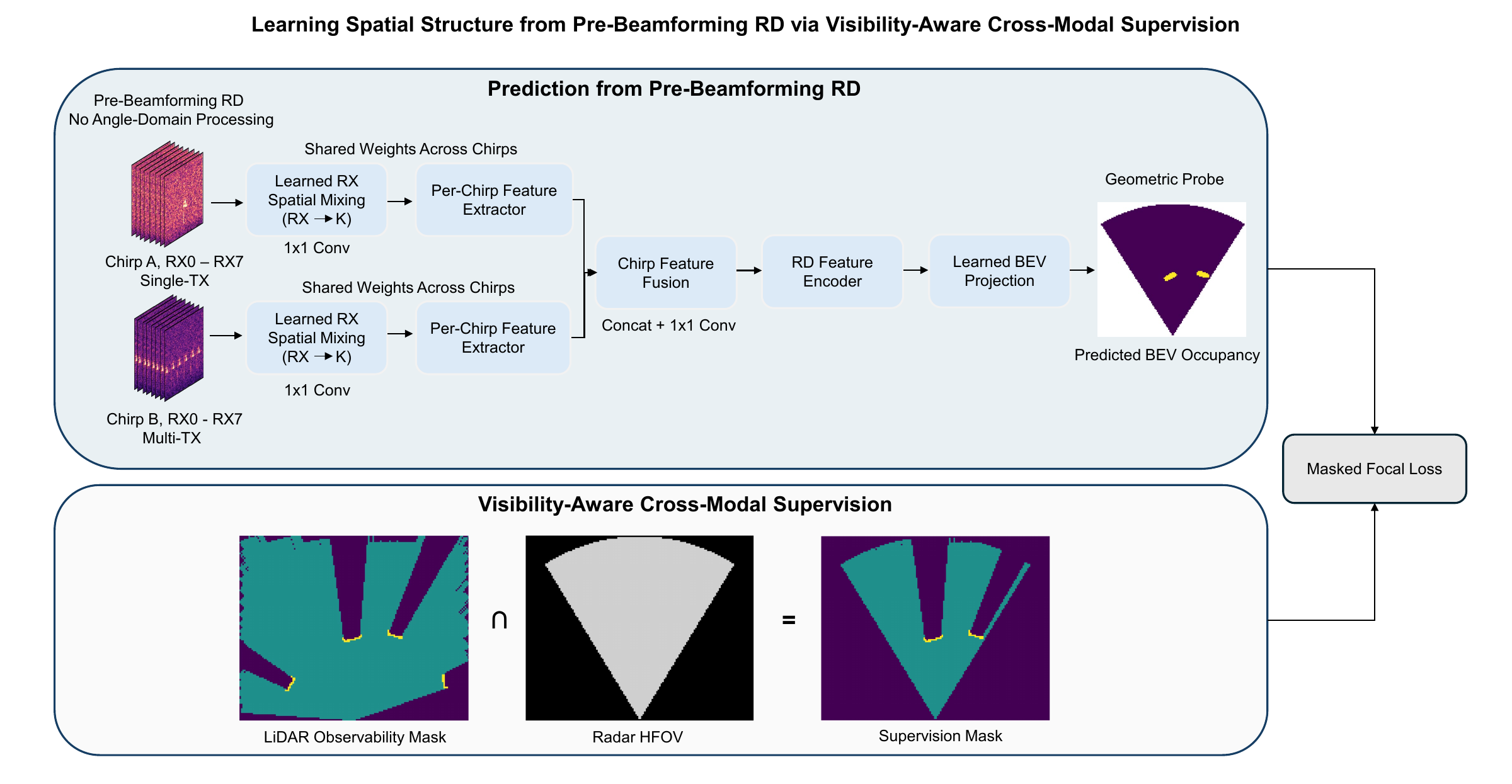}
\caption{
Overview of the proposed approach. \textbf{Top:} Prediction from pre-beamforming per-antenna range-Doppler (RD) tensors using learned spatial mixing and convolutional RD-to-BEV mapping in an end-to-end manner. Each chirp uses measurements from all eight physical receive antennas; differing only in transmit configuration (single-TX vs. multi-TX). \textbf{Bottom:} Visibility-aware cross-modal supervision constructed by intersecting the radar horizontal field-of-view (HFOV) with the LiDAR observability mask. In the supervision mask, LiDAR-observable occupied (yellow) and free (teal) cells within the radar HFOV are retained, while LiDAR-unobservable regions (purple) within the HFOV are treated as unknown and excluded from supervision; the radar HFOV is shown in white (with black denoting outside-HFOV regions in the intermediate mask), and regions outside the HFOV are not considered during training or evaluation. Training uses masked focal loss restricted to the valid region, enabling evaluation of geometric recoverability from RD as a probe task.
}
    \label{fig:pipeline}
\end{figure*}

To probe spatial structure, we use BEV occupancy as a diagnostic task. Supervision is visibility-aware and cross-modal, derived from LiDAR with explicit modeling of radar horizontal field-of-view (HFOV) and occlusion-aware LiDAR observability through ray-based visibility. This restricts training and evaluation to regions with valid LiDAR labels within the radar HFOV, enabling analyses of spatial recoverability. The overall approach is illustrated in Fig.~\ref{fig:pipeline}.

Our contributions are as follows:

\begin{itemize}

\item \textbf{Learning spatial structure from pre-beamforming RD.}
We investigate whether meaningful spatial structure is recoverable from pre-beamforming per-antenna RD tensors through learned spatial mixing using a dual-chirp shared-weight encoder in an end-to-end manner.

\item \textbf{Signal and radar configuration analysis.}
We investigate how signal properties and radar configurations, including transmit configuration (A-only, B-only, and A+B), receive aperture, and receive-channel ordering, influence spatial recoverability.

\item \textbf{BEV occupancy as a geometric probe.}
We use BEV occupancy not as a performance benchmark but as a diagnostic task to evaluate whether spatial geometry can be learned from pre-beamforming RD measurements.

\item \textbf{Visibility-aware cross-modal supervision and physics-aligned evaluation.}
We introduce a LiDAR-based supervision protocol with explicit modeling of radar HFOV and LiDAR label validity, enabling evaluation with unknown-region handling and physics-aligned analyses of geometric recoverability.

\end{itemize}


\section{RELATED WORK}

\subsection{Radar Signal Representations and RD-Based Learning}

Learning-based radar perception spans multiple signal representations depending on the stage of processing~\cite{10952908,11683914}. Many pipelines operate after angle-domain processing using RA maps, RAD tensors, or 4D radar cubes (e.g.,~\cite{9022248,11355718}, RODNet~\cite{9353210}, RADDet~\cite{9469418}, and transformer-based radar models~\cite{Huang_2025_ICCV}). Other approaches operate on post-processed radar outputs such as CFAR-based radar point clouds~\cite{scheiner_object_2021, Fent_2023_CVPR}. Accordingly, we focus on learned radar representations and review prior work that uses learned perception models rather than classical signal-processing pipelines for detection or angle estimation.

RD representations arise at different stages of the radar processing pipeline, either before beamforming as per-antenna RD measurements with implicit angular structure, or after angle-domain processing as angle-resolved RD slices from RAD representations~\cite{10952908}. FFT-RadNet learns a latent RA representation from per-receiver RD inputs for vehicle detection and free-space estimation in high-definition radar systems with 192 virtual antennas~\cite{9879148}. In contrast, our work operates directly on pre-beamforming per-antenna RD measurements acquired under an A/B CS-FMCW waveform with chirp-dependent transmit configurations, enabling controlled analyses of how varying effective transmit apertures influence spatial recoverability. Rather than producing task-specific outputs such as object detections and free-space estimates, we predict dense BEV occupancy as a geometric probe, capturing continuous scene geometry, including the spatial extent of obstacles, terrain, and other extended structures. DAROD performs object detection directly on RD representations (referred to as RD maps in~\cite{9827281}), while DopplerFormer leverages velocity supervision on RD inputs to improve radar-based object detection~\cite{11127345}. RadarMOTR applies transformer-based architectures on RD representations (also referred to as RD maps in~\cite{10994166}) for multi-object tracking. In parallel, several works explore learning directly from raw ADC signals to reduce reliance on handcrafted processing, including CNN-Swin ADC~\cite{11130063} and ADCNet~\cite{ADCNet}.

While these works demonstrate that learning from RD or raw radar signals can support radar perception tasks such as detection, tracking, and free-space estimation, their primary objective is improving downstream perception performance, typically in high-resolution or fixed-MIMO radar regimes. In contrast, our work investigates whether pre-beamforming per-antenna RD tensors contain sufficient information to recover dense spatial structure directly in BEV under the studied A/B CS-FMCW radar configuration.

\subsection{Radar-Based Occupancy and Scene Completion}

Previous work has explored LiDAR-supervised radar occupancy learning from processed radar measurements, including learned inverse sensor models~\cite{8793263} and semantic occupancy learning~\cite{9022091}. More recent work has explored dense radar occupancy and scene completion~\cite{ding_radarocc_2024, 10777549, Huang_2025_ICCV, 10731871}. Methods such as RadarOcc operate on angle-resolved 4D radar tensors~\cite{ding_radarocc_2024}, while LiCROcc uses radar point clouds with cross-modal distillation to improve semantic occupancy performance~\cite{10777549}. Large-scale transformer models trained on 4D radar cubes further show that radar can produce dense BEV and occupancy predictions under fixed MIMO configurations~\cite{Huang_2025_ICCV}. LiDAR-supervised radar occupancy detectors operating on range-azimuth-elevation-Doppler (RAED) cubes have also been proposed (e.g., RaDelft~\cite{10731871}).

These approaches rely on angle-resolved or post-processed radar representations and aim to improve occupancy accuracy. In contrast, we study a different representational stage by operating directly on pre-beamforming per-antenna RD tensors, before explicit angle-domain construction. BEV occupancy is therefore used to evaluate whether spatial structure can emerge from learned cross-antenna mixing of pre-beamforming RD measurements.

\subsection{Public Radar Datasets}

Public radar datasets span multiple representations, including angle-resolved tensors (RADDet~\cite{9469418}, CARRADA~\cite{9413181}, K-Radar~\cite{NEURIPS2022_185fdf62}), post-processed radar point clouds (7V-Scanario~\cite{7V-Scanario_2025}, RadarScenes~\cite{9627037}), and mechanical radar imagery (Oxford Radar RobotCar~\cite{9196884}, RADIATE~\cite{9562089}, Boreas~\cite{Boreas}).

Datasets with lower-level signal access have also emerged. RADIal provides raw ADC recordings and supports generation of RD, RA, and RAD representations from a high-definition MIMO radar (12-TX $\times$ 16-RX)~\cite{9879148}. The I/Q-1M dataset provides large-scale raw I/Q measurements (3-TX $\times$ 4-RX) enabling learning on 4D radar cubes~\cite{Huang_2025_ICCV}. These datasets employ fixed transmit activation patterns within each radar cycle.

In contrast, our work studies pre-beamforming per-antenna RD measurements under the A/B CS-FMCW waveform, in which the transmit antenna activation pattern varies across chirps within a radar cycle (single-TX A-ramp vs.\ multi-TX B-ramp). This produces chirp-dependent effective transmit apertures within a radar cycle, enabling controlled A-only, B-only, and A+B analyses of aperture-dependent spatial recoverability that are not possible with fixed transmit activation patterns. To our knowledge, publicly available automotive radar datasets employ fixed MIMO activation patterns and do not expose chirp-level transmit antenna activation required for controlled aperture analyses within a radar cycle.


\section{Radar Representation and Dataset Setup}

\subsection{Pre-Beamforming Per-Antenna RD Measurements}

The radar sensor provides complex pre-beamforming RD measurements for each receive antenna. For each frame we obtain a tensor:

\begin{equation}
\mathbf{X} \in \mathbb{R}^{C \times N_{\mathrm{rx}} \times R \times D \times 2},
\label{eq:input_tensor}
\end{equation}

where $C$ denotes the two chirp types in the A/B waveform, $R$ the number of range bins, $N_{\mathrm{rx}}$ the number of receive antennas, and $D$ the number of Doppler bins. The final dimension represents the real and imaginary components of the complex signal. In our setup the RD tensor uses $R = 200$ range bins and $D = 128$ Doppler bins, corresponding to a radar range resolution of approximately $0.33\,\mathrm{m}$ per bin. 

The radar employs an A/B chirp-sequence waveform in which the two chirp types activate different transmit antenna configurations (single-TX and multi-TX). The RD tensors are extracted prior to angle-domain processing, preserving the per-antenna phase relationships that encode spatial cues. Since the sensor primarily captures spatial structure in the horizontal (azimuth) plane, with limited elevation resolution, we adopt BEV as the geometric probe task.

\subsection{Sensor Setup and Data Collection}
\label{sec:sensor_setup}

\begin{figure}[t]
\centering
\includegraphics[height=4.8cm]{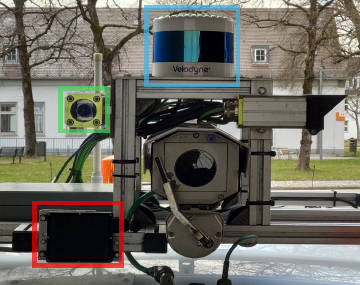}
\caption{Sensor setup with radar (red), LiDAR (blue), and camera (green).}
\label{fig:sensor_setup}
\end{figure}

Data collection was performed using a research vehicle equipped with a Smartmicro~DRVEGRD~152 radar (76-77\,GHz, 6-TX $\times$ 8-RX, 64$^\circ$ HFOV, mid-range mode 65\,m, 18\,Hz) operating in radar cube streaming mode, providing pre-beamforming per-antenna RD tensors without additional on-device detection processing~\cite{smartmicro_drvegrd152}, a Velodyne Alpha Prime LiDAR (128 channels, 10\,Hz), and a Basler acA2440-20gc RGB camera (10\,Hz). The LiDAR is roof-mounted, while the radar is mounted below the LiDAR, as shown in Fig.~\ref{fig:sensor_setup}.

Data were recorded on campus roads and automotive test-track environments containing both stationary and dynamic scenes with ego-motion and moving vehicles, pedestrians, roadside vegetation, and terrain slopes. Radar and LiDAR frames are aligned via nearest-neighbor temporal matching (within a small temporal offset) to account for differing sensor frame rates, and the resulting radar-LiDAR pairs are used for cross-modal supervision.
  
The dataset contains 16,600 synchronized radar-LiDAR frames, split into 11,680 training and 4,920 validation frames ($\approx 70/30$), with evaluation performed on the validation set. Splits are created at the sequence level to avoid temporal overlap between training and validation data.


\section{Methodology}

\subsection{Visibility-Aware Supervision}
\label{sec:visibility_supervision}

LiDAR supervision is generated in BEV using ground-removed point clouds (via a simple height-based filter in the LiDAR frame). The BEV plane is discretized into a grid, and cells containing at least one non-ground LiDAR return are labeled as occupied.

To account for LiDAR occlusions, we compute a BEV observability mask using 2D ray casting from the LiDAR origin over discretized azimuth bins (0.05$^\circ$ resolution), similar to ray-based visibility modeling used in occupancy label generation~\cite{NEURIPS2023_cabfaeec} and visibility-aware radar occupancy learning~\cite{8793263}. The nearest non-ground return acts as an occluder; if no obstacle exists, the farthest LiDAR return defines the free-space extent. Cells along each ray up to the endpoint are marked observable, while cells beyond the endpoint are treated as unobserved.

Supervision is applied only where valid LiDAR labels are available within the radar HFOV. The supervision mask is defined as

\begin{equation}
M_{\mathrm{sup}} = M_{\mathrm{HFOV}} \cap M_{\mathrm{LiDAR\text{-}Obs}},
\label{eq:msup}
\end{equation}

where $M_{\mathrm{HFOV}}$ denotes the radar HFOV mask and $M_{\mathrm{LiDAR\text{-}Obs}}$ denotes the LiDAR observability mask (cells observed by LiDAR, either free or occupied). Intersecting with the radar $M_{\mathrm{HFOV}}$ restricts supervision to regions where valid LiDAR labels exist within the radar sensing sector. Cells within the radar HFOV that are not observable by LiDAR are treated as unknown and excluded from supervision, but are retained for evaluation of unknown-region hallucination. Regions outside the radar HFOV are not considered during training or evaluation (Fig.~\ref{fig:pipeline}, bottom). 

Because LiDAR observability varies with scene geometry and occlusions, the supervision mask is scene-dependent, resulting in a partially supervised learning setting. The supervision mask therefore models the validity of LiDAR supervision rather than radar observability, which additionally depends on radar-specific sensing characteristics, such as antenna pattern, SNR, material properties, and multipath propagation. Accordingly, LiDAR provides a high-resolution geometric reference for supervision, but is used as a geometric proxy rather than radar-equivalent ground truth due to differing sensing physics, occlusion behavior, and sensor mounting offset. Additionally, BEV occupancy exhibits strong class imbalance, with free-space cells significantly outnumbering occupied cells. Training therefore employs a masked focal loss~\cite{8417976} computed only over $i \in M_{\mathrm{sup}}$.

\subsection{Pre-Beamforming RD-to-BEV Learning Network}

The proposed network maps the pre-beamforming per-antenna RD tensor $\mathbf{X}$ defined in Eq.~\eqref{eq:input_tensor} directly to BEV occupancy predictions, as illustrated in Fig.~\ref{fig:pipeline} (top).

The input consists of two RD tensors corresponding to the two chirp types of the A/B waveform. For each chirp, the complex receive-antenna measurements are arranged into a channel-first tensor of size $2N_{\mathrm{rx}} \times R \times D$, where the factor of 2 corresponds to the real and imaginary components, and normalized at each RD cell by the square root of the mean power across receive antennas. Each chirp branch then applies a receive-antenna mixing layer (Rx$\rightarrow$K), implemented as a $1\times1$ convolution that projects the complex receive-channel representation into a latent feature space of dimension $K=64$, while preserving the range-Doppler resolution. The mixing weights are shared across chirp branches, as both chirps produce RD tensors with the same signal structure, differing only in effective transmit aperture. The resulting features are processed by a shared per-chirp feature extractor.

The two chirp feature streams are fused by concatenation followed by a $1\times1$ convolution, producing a joint RD representation. This fused representation is processed by an RD encoder, which reduces resolution while extracting higher-level features.

Finally, the encoded RD representation is rearranged into a compact feature representation and processed by a lightweight U-Net-style convolutional encoder-decoder~\cite{u_net_2015}, which learns the RD-to-BEV projection. The resulting BEV features are further refined by residual convolutional layers before the prediction head outputs occupancy logits. The entire architecture is trained end-to-end, with the RD-to-BEV mapping learned directly from data.


\section{Experiments}

\subsection{Training Setup}

The BEV grid is defined in the LiDAR coordinate frame at 0.5\,m per cell over $x\in[0,60]$\,m and $y\in[-38,38]$\,m, yielding a grid of $H\times W = 120\times152$ cells. Additional experiments evaluate resolutions of 0.4\,m and 0.35\,m over the same spatial extent. Pre-beamforming RD tensors are provided in the radar frame without explicit geometric transformation. The network predicts occupancy in the LiDAR BEV frame via cross-modal supervision, with residual mismatch from differing viewpoints, sensing physics, and sensor offset handled implicitly by the learned mapping.

The network is trained end-to-end on the training split using AdamW~\cite{loshchilov2018decoupled} with an initial learning rate of $10^{-4}$, cosine decay, 50 epochs, and batch size 4. The model contains approximately 3.2M trainable parameters and is trained from scratch.

Supervision is applied only within the visibility-aware mask $M_{\mathrm{sup}}$ defined in Eq.~\eqref{eq:msup}, resulting in partial supervision over cells with valid LiDAR labels within the radar HFOV. Due to the strong free/occupied imbalance, training uses masked focal loss~\cite{8417976} computed only over $i\in M_{\mathrm{sup}}$.

\subsection{Evaluation Protocol}

Evaluation is performed on the validation set within $M_{\mathrm{sup}}$ defined in Eq.~\eqref{eq:msup}. Performance is reported using average precision (AP), defined as the area under the precision-recall curve (AUPRC), computed over pixel-wise BEV occupancy predictions. We additionally report occupied-class Intersection over Union (IoU), as free-space dominates the BEV grid, using a global threshold selected by maximizing the F1 score on the validation set for each model, and applied uniformly across all bands. All metrics are reported in the range [0,1].

To analyze spatial behavior, performance is reported across range bands (0-20\,m, 20-40\,m, 40-60\,m) and angular sectors within the radar HFOV. Predictions outside the LiDAR observability mask are excluded from both AP and IoU. Unknown-region behavior is evaluated separately using the unknown-region hallucination rate (UHR), defined as the fraction of predicted occupied cells within LiDAR-unobservable regions ($M_{\mathrm{HFOV}} \cap \neg M_{\mathrm{LiDAR\text{-}Obs}}$) at the model's global validation threshold.

We include two simple radar baselines for reference:

\begin{itemize}
\item \textbf{Random prior}, predicting a constant occupancy probability equal to the empirical fraction of occupied cells within the supervised BEV region $M_{\mathrm{sup}}$, estimated from the dataset distribution.

\item \textbf{Range-energy projection}, a physics-inspired baseline obtained by averaging RD magnitude across chirps, antennas, and Doppler bins to produce a normalized 1D range energy profile, which is then mapped to BEV cells based on their radial distance, effectively assuming azimuthal symmetry to form a coarse occupancy estimate.

\end{itemize}

Reliable conventional beamforming requires accurate sensor-specific antenna calibration to compensate for hardware-induced inter-channel phase and gain offsets. As the factory calibration parameters required for such reconstruction are proprietary and unavailable for the employed radar platform, a scientifically validated conventional beamforming baseline could not be established.

\subsection{Main Results}

\begin{table}[t]
\centering
\caption{Main results at 0.5\,m BEV resolution.}
\label{tab:main_results}
\begin{tabular}{lccc}
\toprule
\textbf{Method} & \textbf{AP} $\uparrow$ & \textbf{IoU} $\uparrow$ & \textbf{UHR} $\downarrow$ \\
\midrule
Random prior & 0.05 & -- & -- \\
Range-energy projection & 0.06 & 0.06 & 0.17 \\
\textbf{Ours} & \textbf{0.36} & \textbf{0.24} & \textbf{0.11} \\
\bottomrule
\end{tabular}
\end{table}

\begin{figure}[t]
\centering
\includegraphics[width=\columnwidth]{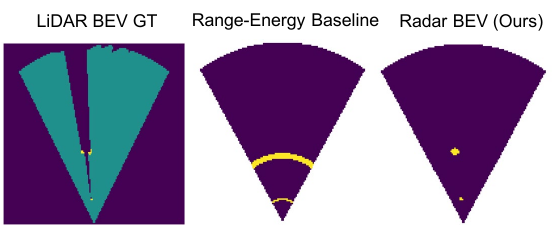}
\caption{Comparison between LiDAR BEV GT, the range-energy (radial) baseline, and the proposed method. The baseline produces radially symmetric responses due to the absence of angular information, while the proposed model recovers spatially localized structure from pre-beamforming RD.}
\label{fig:baseline_comparison}
\end{figure}

Table~\ref{tab:main_results} reports performance at the base BEV resolution of 0.5\,m. 
The proposed method substantially outperforms both baselines, achieving a large improvement over the random prior and the range-energy projection baseline. The range-energy baseline provides marginal improvement over the random prior, indicating that range-only aggregation provides limited geometric structure in the absence of angular discrimination. This behavior is illustrated qualitatively in Fig.~\ref{fig:baseline_comparison}. We additionally evaluated non-shared chirp-branch weights and observed comparable performance (AP 0.35 vs.\ 0.36, IoU 0.24 for both); shared chirp-branch weights are therefore retained for parameter efficiency.

Overall, the learned model achieves significantly higher AP and IoU, demonstrating that spatial structure can be recovered directly from pre-beamforming per-antenna RD measurements through learned spatial mixing. The proposed method reduces hallucination in unknown regions compared to the range-energy baseline, indicating more reliable spatial reasoning beyond observed areas.

The absolute performance reflects the inherent difficulty of cross-modal BEV occupancy prediction between radar and LiDAR due to their differing sensing physics and the partial observability imposed by the supervision mask. Consequently, IoU should be interpreted together with the qualitative results rather than as an absolute measure of geometric reconstruction fidelity.

\subsection{Qualitative Results}

\begin{figure}[t]
\centering
\includegraphics[width=\columnwidth]{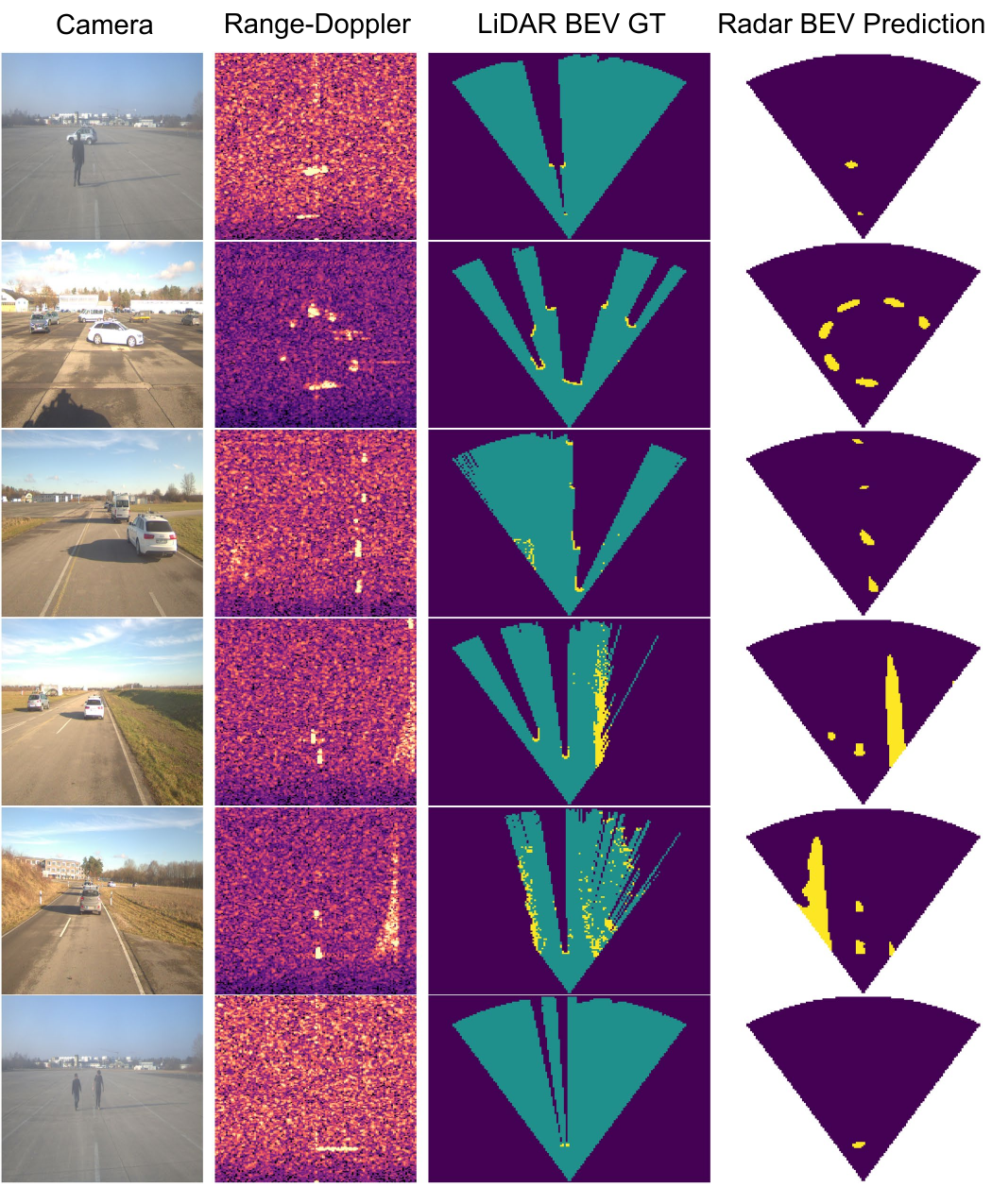}
\caption{Qualitative radar BEV occupancy predictions from pre-beamforming RD tensors. Each row shows (from left to right) the camera image (for visualization only), RD magnitude (single RX, chirp A; range on the vertical axis, Doppler on the horizontal axis, zero Doppler centered), LiDAR BEV ground truth within the visible region, and the radar BEV prediction. For visualization, RD is shown for a single receive channel and chirp with Doppler centered via FFT shift and magnitude log-compressed and normalized, while the model operates on all receive channels and both chirps (A+B) using the native RD representation without these visualization transformations. In the LiDAR BEV ground-truth (GT) panel, occupied cells are shown in yellow, free space in teal, and LiDAR-unobservable regions within the radar HFOV in purple (unknown and excluded from supervision), while regions outside the radar HFOV are not considered during training or evaluation; in the radar BEV prediction, non-occupied cells are shown in purple. Representative successful predictions and failure cases.}
\label{fig:qualitative}
\end{figure}

Fig.~\ref{fig:qualitative} presents representative examples. Large structures such as vehicles and extended terrain (e.g., slopes) are generally recovered with coherent occupancy responses.

Predictions are spatially more diffuse than LiDAR ground truth, reflecting the limited angular resolution and speckle characteristics of automotive radar. This leads to blob-like responses and slight spatial offsets relative to LiDAR annotations, which can reduce pixel-wise agreement despite consistent object-level structure.

Failure cases are observed for small or closely spaced objects, where limited angular resolution and multipath effects lead to merged or ambiguous responses. Differences relative to LiDAR also arise from cross-modal misalignment and sensing physics; notably, radar can respond in partially occluded regions not observed by LiDAR, reflecting complementary sensing rather than purely erroneous predictions. These observations suggest that the learned representation captures radar-specific phenomena beyond those directly represented in the LiDAR supervision.

The qualitative results are consistent with the quantitative evaluation: predictions are less spatially precise than the LiDAR reference, but still capture coherent scene structure, supporting the use of BEV occupancy as a probe rather than as a reconstruction of LiDAR measurements.

\subsection{Signal Representation Analysis}

To better understand which information in the complex pre-beamforming RD signal contributes to spatial recoverability, Table~\ref{tab:signal_representation} analyzes signal-component and phase-perturbation variants. Signal-component variants are trained under the corresponding input representation. For the phase-scramble variant, phase values are randomly permuted across receive channels independently at each chirp, range, and Doppler cell while preserving the corresponding magnitudes. A different random permutation is generated for each sample and kept fixed throughout training and evaluation. To isolate the effect of absolute phase, global phase perturbations are evaluated only at inference using the model trained on the full input representation.

Magnitude-only input leads to a substantial performance degradation, whereas phase-only input retains most of the full-model performance. Together with the degradation under phase scrambling, these results indicate that channel-consistent inter-antenna phase relationships provide the primary contribution to spatial recoverability, while magnitude provides complementary information. Applying a common global phase shift of $90^\circ$ and $180^\circ$ to the full complex RD tensor at inference does not affect performance, indicating that the learned representation depends primarily on relative inter-antenna phase rather than absolute phase.

\begin{table}[t]
\centering
\caption{Effect of signal representation on spatial recoverability.}
\label{tab:signal_representation}
\small
\setlength{\tabcolsep}{4pt}
\renewcommand{\arraystretch}{0.95}
\begin{tabular}{llcc}
\toprule
\textbf{Property} & \textbf{Variant} &
\textbf{AP} $\uparrow$ & \textbf{IoU} $\uparrow$ \\
\midrule

\multirow{4}{*}{Signal component}
& \textbf{Full}            & \textbf{0.36} & \textbf{0.24} \\
& Magnitude only           & 0.17 & 0.13 \\
& Phase only               & 0.33 & 0.22 \\
& Phase scramble           & 0.19 & 0.14 \\

\midrule

\multirow{1}{*}{\shortstack[l]{Global phase}}
& Shift ($90^\circ$, $180^\circ$) & 0.36 & 0.24 \\

\bottomrule
\end{tabular}
\end{table}

\subsection{Radar Configuration Analysis}

\begin{table}[t]
\centering
\caption{Effect of radar configuration on spatial recoverability.}
\label{tab:signal_properties_chirp_aperture_rx_ordering}
\small
\setlength{\tabcolsep}{3pt}
\renewcommand{\arraystretch}{0.95}

\begin{tabular}{@{}l cc cc cc@{}}
\toprule
\multirow[c]{2}{*}{\textbf{RX aperture}}
& \multicolumn{2}{c}{\textbf{A+B}}
& \multicolumn{2}{c}{\textbf{B-only}}
& \multicolumn{2}{c}{\textbf{A-only}} \\
\cmidrule(lr){2-3}
\cmidrule(lr){4-5}
\cmidrule(lr){6-7}
& \textbf{AP} $\uparrow$ & \textbf{IoU} $\uparrow$
& \textbf{AP} $\uparrow$ & \textbf{IoU} $\uparrow$
& \textbf{AP} $\uparrow$ & \textbf{IoU} $\uparrow$ \\
\midrule
\textbf{Full}
& \textbf{0.36} & \textbf{0.24}
& 0.34 & 0.23
& 0.28 & 0.19 \\
Half
& 0.24 & 0.17
& 0.20 & 0.14
& 0.13 & 0.11 \\
Quarter
& 0.15 & 0.12
& 0.11 & 0.10
& 0.08 & 0.08 \\
\midrule
\multicolumn{3}{@{}l}{\textbf{RX ordering}}
& \multicolumn{2}{c}{\textbf{AP} $\uparrow$}
& \multicolumn{2}{c@{}}{\textbf{IoU} $\uparrow$} \\
\cmidrule(lr){1-3}
\cmidrule(lr){4-5}
\cmidrule(l){6-7}
\multicolumn{3}{@{}l}{\textbf{Original}}
& \multicolumn{2}{c}{\textbf{0.36}}
& \multicolumn{2}{c@{}}{\textbf{0.24}} \\
\multicolumn{3}{@{}l}{Fixed reorder}
& \multicolumn{2}{c}{0.34}
& \multicolumn{2}{c@{}}{0.23} \\
\multicolumn{3}{@{}l}{Random reorder}
& \multicolumn{2}{c}{0.21}
& \multicolumn{2}{c@{}}{0.15} \\
\bottomrule
\end{tabular}
\end{table}

\begin{figure}[t]
\centering
\includegraphics[width=\columnwidth]{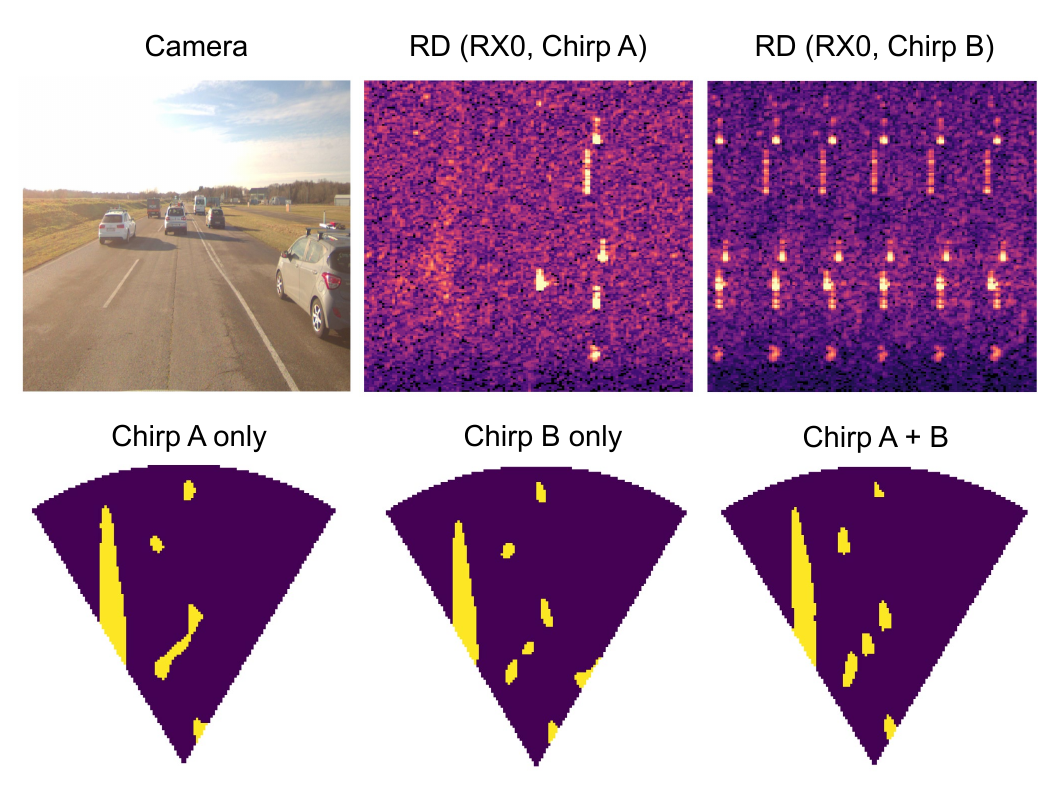}
\caption{Qualitative comparison of transmit configurations for a representative scene. Top row shows the camera view and RD magnitude (single RX) for Chirp A and Chirp B (shown for visualization; the model uses all receive channels for each chirp configuration). Bottom row shows BEV predictions using Chirp~A, Chirp~B, and both chirps (A+B). The visual progression is consistent with the quantitative results in Table~\ref{tab:signal_properties_chirp_aperture_rx_ordering}.} 
\label{fig:chirp_analysis}
\end{figure}

Table~\ref{tab:signal_properties_chirp_aperture_rx_ordering} analyzes how radar configuration influences spatial recoverability by varying transmit configuration, receive aperture, and receive-channel ordering. All variants are trained under the corresponding configuration while keeping the network architecture unchanged. For A-only and B-only variants, the absent chirp input is zeroed while the chirp-fusion module remains unchanged. Receive aperture variants are constructed by zeroing inactive receive channels while preserving the original tensor shape. For RX ordering, the network is trained and evaluated using either the original ordering, a fixed alternative ordering, or a random ordering generated independently for each sample.

At full receive aperture, B-only outperforms A-only, consistent with the larger effective transmit aperture of the multi-TX configuration. Combining both chirp types (A+B) achieves the highest performance, suggesting complementary information for spatial recoverability (Fig.~\ref{fig:chirp_analysis}). Because chirps A and B also differ in other chirp-dependent characteristics (e.g., SNR, timing, and Doppler coupling), we interpret these results as aperture-consistent rather than attributing the observed gains solely to aperture. Performance also decreases progressively as the receive aperture is reduced, consistent with the importance of receive aperture for spatial recoverability.

RX ordering experiments further indicate that preserving a consistent receive-channel ordering is important for spatial recoverability. The small performance drop when training and evaluating with a fixed alternative channel ordering suggests that the learned representation can still accommodate a different but consistent receive-channel ordering. In contrast, using a random channel ordering per sample causes a substantially larger degradation, suggesting that the learned representation exploits stable inter-channel relationships.


\subsection{Range and Doppler Analysis}

To assess the contribution of range and Doppler information, we evaluate collapsed RD variants summarized in Table~\ref{tab:rd_ablation}. Collapsed RD variants are trained under the corresponding input representation. Doppler- and range-collapsed variants are constructed by averaging over the corresponding RD axis and broadcasting the result back to the original tensor shape while keeping the network architecture unchanged.

Collapsing the Doppler dimension results in a moderate performance degradation, indicating that Doppler provides complementary information beyond range alone. In contrast, collapsing the range dimension causes a severe performance drop and a substantial increase in hallucination, highlighting the dominant role of range for spatial localization.

\begin{table}[!t]
\centering
\caption{Effect of range and Doppler information.}
\label{tab:rd_ablation}
\begin{tabular}{lccc}
\toprule
\textbf{Variant} & \textbf{AP} $\uparrow$ & \textbf{IoU} $\uparrow$ & \textbf{UHR} $\downarrow$ \\
\midrule
\textbf{Full RD} & \textbf{0.36} & \textbf{0.24} & \textbf{0.11} \\
Doppler-collapsed & 0.29 & 0.20 & 0.12 \\
Range-collapsed & 0.08 & 0.07 & 0.25 \\
\bottomrule
\end{tabular}
\end{table}

\subsection{BEV Resolution Study}

\begin{table}[!t]
\centering
\caption{Effect of BEV resolution on performance.}
\label{tab:resolution}
\begin{tabular}{lcc}
\toprule
\textbf{Resolution} & \textbf{AP} $\uparrow$ & \textbf{IoU} $\uparrow$ \\
\midrule
\textbf{0.5\,m} & \textbf{0.36} & \textbf{0.24} \\
0.4\,m & 0.30 & 0.21 \\
0.35\,m & 0.27 & 0.19 \\
\bottomrule
\end{tabular}
\end{table}

As shown in Table~\ref{tab:resolution}, performance decreases as the BEV grid resolution becomes finer, with both AP and IoU dropping from 0.5\,m to 0.35\,m.

This trend reflects the increased localization difficulty at finer resolutions, where smaller cells require more precise spatial predictions. Finer grids also reduce the fraction of occupied cells, increasing class imbalance, while the limited angular resolution and diffuse spatial responses of automotive radar make accurate cell-level localization more challenging. Nevertheless, the model retains non-trivial performance across all evaluated resolutions, with 0.5\,m providing the best balance between spatial resolution and recoverability.

\subsection{Band-wise Analysis}

\begin{table}[!t]
\centering
\caption{Band-wise performance at 0.5\,m BEV resolution.}
\label{tab:bandwise}
\begin{tabular}{lccc}
\toprule
\textbf{Band} & \textbf{AP} $\uparrow$ & \textbf{IoU} $\uparrow$ & \textbf{pos\_frac } \\
\midrule
Overall & 0.36 & 0.24 & 0.05 \\
\midrule
0--20\,m & 0.44 & 0.29 & 0.06 \\
20--40\,m & 0.38 & 0.25 & 0.06 \\
40--60\,m & 0.25 & 0.19 & 0.04 \\
\midrule
Center (0--15$^\circ$) & 0.33 & 0.23 & 0.05 \\
Edges (15--32$^\circ$) & 0.38 & 0.25 & 0.06 \\
\bottomrule
\end{tabular}
\end{table}


Performance across range and angular bands is summarized in Table~\ref{tab:bandwise}. AP decreases with increasing range, from 0--20\,m to 40--60\,m, consistent with reduced signal strength and increased sparsity of radar returns. Lower \textit{pos\_frac} in the far range is associated with greater class imbalance, which may further contribute to the observed degradation.

Across angular bands, performance is slightly higher in the edge regions than in the center, although the difference is modest. This variation may reflect differences in scene structure and radar coverage across the field of view.


\section{Conclusion}

This work investigates spatial recoverability directly from pre-beamforming per-antenna RD measurements using learned spatial mixing, with BEV occupancy as a geometric probe under visibility-aware cross-modal supervision.

Experimental results show that the learned model substantially outperforms simple radar baselines while signal-property, radar-configuration, and range-Doppler analyses provide consistent evidence for the factors influencing spatial recoverability. Analyses further indicate that geometric recoverability depends on effective transmit aperture and preservation of inter-antenna signal structure, while degrading with increasing spatial resolution and range.

These findings indicate that meaningful spatial structure is recoverable from pre-beamforming per-antenna RD measurements under the studied A/B CS-FMCW radar configuration. The proposed framework therefore provides a representation probe for studying spatial recoverability through learned spatial mixing, rather than a physically interpretable replacement for conventional angle-domain processing.


\balance  

                                  


\bibliographystyle{IEEEtran}
\bibliography{root_citations}  

@INPROCEEDINGS{9879148,
  author={Rebut, Julien and Ouaknine, Arthur and Malik, Waqas and Pérez, Patrick},
  booktitle={2022 IEEE/CVF Conference on Computer Vision and Pattern Recognition (CVPR)}, 
  title={{Raw High-Definition Radar for Multi-Task Learning}}, 
  year={2022},
  volume={},
  number={},
  pages={},
  doi={10.1109/CVPR52688.2022.01651}}

@InProceedings{Huang_2025_ICCV,
    author    = {Huang, Tianshu and Prabhakara, Akarsh and Chen, Chuhan and Karhade, Jay and Ramanan, Deva and O'toole, Matthew and Rowe, Anthony},
    title     = {{Towards Foundational Models for Single-Chip Radar}},
    booktitle = {Proceedings of the IEEE/CVF International Conference on Computer Vision (ICCV)},
    month     = {},
    year      = {2025},
    pages     = {24655--24665}
}

@INPROCEEDINGS{9469418,
  author={Zhang, Ao and Nowruzi, Farzan Erlik and Laganiere, Robert},
  booktitle={2021 18th Conference on Robots and Vision (CRV)}, 
  title={{RADDet: Range-Azimuth-Doppler based Radar Object Detection for Dynamic Road Users}}, 
  year={2021},
  volume={},
  number={},
  pages={},
  doi={10.1109/CRV52889.2021.00021}}

@INPROCEEDINGS{9413181,
  author={Ouaknine, Arthur and Newson, Alasdair and Rebut, Julien and Tupin, Florence and Pérez, Patrick},
  booktitle={Proceedings of the 25th International Conference on Pattern Recognition (ICPR)}, 
  title={{CARRADA Dataset: Camera and Automotive Radar with Range- Angle- Doppler Annotations}}, 
  year={2021},
  volume={},
  number={},
  pages={5068--5075},
  doi={10.1109/ICPR48806.2021.9413181}}

@inproceedings{NEURIPS2022_185fdf62,
 author = {Paek, Dong-Hee and KONG, SEUNG-HYUN and Wijaya, Kevin Tirta},
 booktitle = {Advances in Neural Information Processing Systems},
 pages = {3819--3829},
 publisher = {Curran Associates, Inc.},
 title = {{K-Radar: 4D Radar Object Detection for Autonomous Driving in Various Weather Conditions}},
 url = {},
 volume = {35},
 year = {2022}
}

@InProceedings{7V-Scanario_2025,
  author    = {Philipp Berthold AND Bianca Forkel AND Mirko Maehlisch},
  booktitle = {Symposium Sensor Data Fusion (SDF)},
  title     = {{A Multi-Vehicle Dataset with Camera, LiDAR, and Radar Sensors and Scanned 3D Models for Custom Auto-Annotation using RTK-GNSS}},
  year      = {2025}
}

@INPROCEEDINGS{9627037,
  author={Schumann, Ole and Hahn, Markus and Scheiner, Nicolas and Weishaupt, Fabio and Tilly, Julius F. and Dickmann, Jürgen and Wöhler, Christian},
  booktitle={2021 IEEE 24th International Conference on Information Fusion (FUSION)}, 
  title={{RadarScenes: A Real-World Radar Point Cloud Data Set for Automotive Applications}}, 
  year={2021},
  volume={},
  number={},
  pages={},
  doi={10.23919/FUSION49465.2021.9627037}}

@INPROCEEDINGS{9196884,
  author={Barnes, Dan and Gadd, Matthew and Murcutt, Paul and Newman, Paul and Posner, Ingmar},
  booktitle={2020 IEEE International Conference on Robotics and Automation (ICRA)}, 
  title={{The Oxford Radar RobotCar Dataset: A Radar Extension to the Oxford RobotCar Dataset}}, 
  year={2020},
  volume={},
  number={},
  pages={6433--6438},
  doi={10.1109/ICRA40945.2020.9196884}}

@INPROCEEDINGS{9562089,
  author={Sheeny, Marcel and De Pellegrin, Emanuele and Mukherjee, Saptarshi and Ahrabian, Alireza and Wang, Sen and Wallace, Andrew},
  booktitle={2021 IEEE International Conference on Robotics and Automation (ICRA)}, 
  title={{RADIATE: A Radar Dataset for Automotive Perception in Bad Weather}}, 
  year={2021},
  volume={},
  number={},
  pages={},
  doi={10.1109/ICRA48506.2021.9562089}}

@article{Boreas,
author = {Keenan Burnett and David J Yoon and Yuchen Wu and Andrew Z Li and Haowei Zhang and Shichen Lu and Jingxing Qian and Wei-Kang Tseng and Andrew Lambert and Keith YK Leung and Angela P Schoellig and Timothy D Barfoot},
title ={{Boreas: A Multi-Season Autonomous Driving Dataset}},
journal = {The International Journal of Robotics Research},
volume = {},
number = {},
pages = {},
year = {2023},
doi = {10.1177/02783649231160195},
URL = {
},
eprint = { https://doi.org/10.1177/02783649231160195}
}

@INPROCEEDINGS{8793263,
  author={Weston, Rob and Cen, Sarah and Newman, Paul and Posner, Ingmar},
  booktitle={2019 International Conference on Robotics and Automation (ICRA)}, 
  title={{Probably Unknown: Deep Inverse Sensor Modelling Radar}}, 
  year={2019},
  volume={},
  number={},
  pages={5446--5452},
  doi={10.1109/ICRA.2019.8793263}}

@INPROCEEDINGS{9022091,
  author={Sless, Liat and Shlomo, Bat El and Cohen, Gilad and Oron, Shaul},
  booktitle={2019 IEEE/CVF International Conference on Computer Vision Workshop (ICCVW)}, 
  title={{Road Scene Understanding by Occupancy Grid Learning from Sparse Radar Clusters using Semantic Segmentation}}, 
  year={2019},
  volume={},
  number={},
  pages={},
  doi={10.1109/ICCVW.2019.00115}}

@inproceedings{ding_radarocc_2024,
	title = {{RadarOcc}: {Robust} {3D} {Occupancy} {Prediction} with {4D} {Imaging} {Radar}},
	volume = {37},
	url = {},
	doi = {10.52202/079017-3222},
	booktitle = {Advances in {Neural} {Information} {Processing} {Systems}},
	publisher = {Curran Associates, Inc.},
	author = {Ding, Fangqiang and Wen, Xiangyu and Zhu, Yunzhou and Li, Yiming and Lu, Chris Xiaoxuan},
	year = {2024},
	pages = {101589--101617},
}

@ARTICLE{10777549,
  author={Ma, Yukai and Mei, Jianbiao and Yang, Xuemeng and Wen, Licheng and Xu, Weihua and Zhang, Jiangning and Zuo, Xingxing and Shi, Botian and Liu, Yong},
  journal={IEEE Robotics and Automation Letters}, 
  title={{LiCROcc: Teach Radar for Accurate Semantic Occupancy Prediction Using LiDAR and Camera}}, 
  year={2025},
  volume={10},
  number={1},
  pages={852--859},
  doi={10.1109/LRA.2024.3511427}}

@ARTICLE{10731871,
  author={Roldan, Ignacio and Palffy, Andras and Kooij, Julian F. P. and Gavrila, Dariu M. and Fioranelli, Francesco and Yarovoy, Alexander},
  journal={IEEE Transactions on Radar Systems}, 
  title={{A Deep Automotive Radar Detector Using the RaDelft Dataset}}, 
  year={2024},
  volume={2},
  number={},
  pages={},
  doi={10.1109/TRS.2024.3485578}}

@INPROCEEDINGS{9022248,
  author={Major, Bence and Fontijne, Daniel and Ansari, Amin and Sukhavasi, Ravi Teja and Gowaikar, Radhika and Hamilton, Michael and Lee, Sean and Grzechnik, Slawomir and Subramanian, Sundar},
  booktitle={2019 IEEE/CVF International Conference on Computer Vision Workshop (ICCVW)}, 
  title={{Vehicle Detection With Automotive Radar Using Deep Learning on Range-Azimuth-Doppler Tensors}}, 
  year={2019},
  volume={},
  number={},
  pages={924--932},
  doi={10.1109/ICCVW.2019.00121}}

@ARTICLE{9353210,
  author={Wang, Yizhou and Jiang, Zhongyu and Li, Yudong and Hwang, Jenq-Neng and Xing, Guanbin and Liu, Hui},
  journal={IEEE Journal of Selected Topics in Signal Processing}, 
  title={{RODNet: A Real-Time Radar Object Detection Network Cross-Supervised by Camera-Radar Fused Object 3D Localization}}, 
  year={2021},
  volume={15},
  number={4},
  pages={},
  doi={10.1109/JSTSP.2021.3058895}}

@INPROCEEDINGS{9827281,
  author={Decourt, Colin and VanRullen, Rufin and Salle, Didier and Oberlin, Thomas},
  booktitle={2022 IEEE Intelligent Vehicles Symposium (IV)}, 
  title={{DAROD: A Deep Automotive Radar Object Detector on Range-Doppler maps}}, 
  year={2022},
  volume={},
  number={},
  pages={112--118},
  doi={10.1109/IV51971.2022.9827281}}

@INPROCEEDINGS{11127345,
  author={Zhao, Shuo and Sun, Wei and Li, Huadong and Jiang, Zhaoying},
  booktitle={2025 IEEE International Conference on Robotics and Automation (ICRA)}, 
  title={{Doppler Former: Velocity Supervision of Raw Radar Data}}, 
  year={2025},
  volume={},
  number={},
  pages={13036--13042},
  doi={10.1109/ICRA55743.2025.11127345}}

@INPROCEEDINGS{10994166,
  author={Dell, Martin and Bradfisch, Wolfgang and Schober, Steffen and Klöck, Clemens},
  booktitle={2024 International Radar Conference (RADAR)}, 
  title={{RadarMOTR: Multi-Object Tracking with Transformers on Range-Doppler Maps}}, 
  year={2024},
  volume={},
  number={},
  pages={},
  doi={10.1109/RADAR58436.2024.10994166}}

@INPROCEEDINGS{11130063,
  author={Banwait, Iqbal and Zeller, Niclas and Alirezaie, Javad},
  booktitle={2025 21st International Conference on Intelligent Environments (IE)}, 
  title={{CNN-Swin Backbones in Radar Object Detection for Autonomous Vehicles using Raw ADC Signals}}, 
  year={2025},
  volume={},
  number={},
  pages={},
  doi={10.1109/IE64880.2025.11130063}}

@misc{ADCNet,
      title={{ADCNet: Learning from Raw Radar Data via Distillation}}, 
      author={Bo Yang and Ishan Khatri and Michael Happold and Chulong Chen},
      year={2023},
      eprint={2303.11420},
      archivePrefix={arXiv},
      primaryClass={eess.SP},
      url={https://arxiv.org/abs/2303.11420}, 
}

@article{scheiner_object_2021,
	title = {Object detection for automotive radar point clouds – a comparison},
	volume = {3},
	issn = {2523-398X},
	url = {},
	doi = {10.1186/s42467-021-00012-z},
	number = {1},
	journal = {AI Perspectives},
	author = {Scheiner, Nicolas and Kraus, Florian and Appenrodt, Nils and Dickmann, Jürgen and Sick, Bernhard},
	month = nov,
	year = {2021},
	pages = {},
}

@InProceedings{Fent_2023_CVPR,
    author    = {Fent, Felix and Bauerschmidt, Philipp and Lienkamp, Markus},
    title     = {{RadarGNN: Transformation Invariant Graph Neural Network for Radar-Based Perception}},
    booktitle = {Proceedings of the IEEE/CVF Conference on Computer Vision and Pattern Recognition (CVPR) Workshops},
    month     = {},
    year      = {2023},
    pages     = {182--191}
}

@ARTICLE{7870764,
  author={Patole, Sujeet Milind and Torlak, Murat and Wang, Dan and Ali, Murtaza},
  journal={IEEE Signal Processing Magazine}, 
  title={{Automotive radars: A review of signal processing techniques}}, 
  year={2017},
  volume={34},
  number={2},
  pages={22--35},
  doi={10.1109/MSP.2016.2628914}}

@ARTICLE{8830483,
  author={Saponara, Sergio and Greco, Maria Sabrina and Gini, Fulvio},
  journal={IEEE Signal Processing Magazine}, 
  title={{Radar-on-Chip/in-Package in Autonomous Driving Vehicles and Intelligent Transport Systems: Opportunities and Challenges}}, 
  year={2019},
  volume={36},
  number={5},
  pages={71--84},
  doi={10.1109/MSP.2019.2909074}}

@ARTICLE{10952908,
  author={Yao, Shanliang and Guan, Runwei and Peng, Zitian and Xu, Chenhang and Shi, Yilu and Ding, Weiping and Gee Lim, Eng and Yue, Yong and Seo, Hyungjoon and Lok Man, Ka and Ma, Jieming and Zhu, Xiaohui and Yue, Yutao},
  journal={IEEE Transactions on Intelligent Transportation Systems}, 
  title={{Exploring Radar Data Representations in Autonomous Driving: A Comprehensive Review}}, 
  year={2025},
  volume={26},
  number={6},
  pages={7401--7425},
  doi={10.1109/TITS.2025.3554781}}

@misc{smartmicro_drvegrd152,
  author = {{smartmicro GmbH}},
  title  = {{DRVEGRD 152 RCS Automotive Radar Sensor Datasheet}},
  year   = {2024},
  url    = {https://www.smartmicro.com/wp-content/uploads/2026/03/DRVEGRD_152_Automotive_Data_Sheet.pdf}
}

@inproceedings{
loshchilov2018decoupled,
title={{Decoupled Weight Decay Regularization}},
author={Ilya Loshchilov and Frank Hutter},
booktitle={International Conference on Learning Representations},
year={2019},
url={},
}

@ARTICLE{8417976,
  author={Lin, Tsung-Yi and Goyal, Priya and Girshick, Ross and He, Kaiming and Dollár, Piotr},
  journal={IEEE Transactions on Pattern Analysis and Machine Intelligence}, 
  title={{Focal Loss for Dense Object Detection}}, 
  year={2020},
  volume={42},
  number={2},
  pages={318--327},
  doi={10.1109/TPAMI.2018.2858826}}

@inproceedings{NEURIPS2023_cabfaeec,
 author = {Tian, Xiaoyu and Jiang, Tao and Yun, Longfei and Mao, Yucheng and Yang, Huitong and Wang, Yue and Wang, Yilun and Zhao, Hang},
 booktitle = {Advances in Neural Information Processing Systems},
 pages = {},
 publisher = {Curran Associates, Inc.},
 title = {{Occ3D: A Large-Scale 3D Occupancy Prediction Benchmark for Autonomous Driving}},
 url = {},
 volume = {36},
 year = {2023}
}

@inproceedings{u_net_2015,
	address = {Cham},
	title = {U-{Net}: {Convolutional} {Networks} for {Biomedical} {Image} {Segmentation}},
	isbn = {978-3-319-24574-4},
	booktitle = {Medical {Image} {Computing} and {Computer}-{Assisted} {Intervention} – {MICCAI} 2015},
	publisher = {Springer International Publishing},
	author = {Ronneberger, Olaf and Fischer, Philipp and Brox, Thomas},
	year = {2015},
	pages = {234--241},
}

@ARTICLE{11654526,
  author={Dorvash, Masoud and Alaee-Kerahroodi, Mohammad and Mysore, Bhavani Shankar and Ottersten, Björn and Waldschmidt, Christian and Swindlehurst, A. Lee and Feger, Reinhard},
  journal={Proceedings of the IEEE}, 
  title={{Recent Advances in Millimeter-Wave 4-D Imaging Radars: A Leap Toward Massive MIMO in Sensing}}, 
  year={2026},
  volume={114},
  number={5},
  doi={10.1109/JPROC.2026.3719796}}

@ARTICLE{11683914,
  author={Paek, Dong-Hee and Kong, Seung-Hyun},
  journal={IEEE Intelligent Transportation Systems Magazine}, 
  title={{Artificial Intelligence for Perception Using 4D Radar}}, 
  year={2026},
  volume={},
  number={},
  doi={10.1109/MITS.2026.3727096}}

@ARTICLE{11355718,
  author={Song, Seung-Hyun and Paek, Dong-Hee and Dao, Minh-Quan and Malis, Ezio and Kong, Seung-Hyun},
  journal={IEEE Sensors Journal}, 
  title={{Enhanced 3-D Object Detection via Diverse Feature Representations of 4-D Radar Tensor}}, 
  year={2026},
  volume={26},
  number={5},
  doi={10.1109/JSEN.2026.3652543}}


\end{document}